\pdfoutput=1

\documentclass[11pt]{article}

\usepackage[preprint]{acl}

\usepackage{times}
\usepackage{latexsym}

\usepackage[T1]{fontenc}

\usepackage[utf8]{inputenc}

\usepackage{microtype}

\usepackage{inconsolata}

\usepackage{graphicx}

\usepackage{amsmath}
\usepackage{float}

\title{Probing the contents of semantic representations from text, behavior, and brain data using the \textit{psychNorms} metabase}

\author{
 \textbf{Zak Hussain\textsuperscript{1,2,3}},
 \textbf{Rui Mata\textsuperscript{1}},
 \textbf{Ben R. Newell\textsuperscript{3}},
 \textbf{Dirk U. Wulff\textsuperscript{2,1}},
\\
\\
 \textsuperscript{1}Faculty of Psychology, University of Basel, Basel, Switzerland \\
 \textsuperscript{2}Center for Adaptive Rationality, Max Planck Institute for Human Development, Berlin, Germany \\
 \textsuperscript{3}School of Psychology, UNSW, Sydney, Australia,
\\
 \small{
   \textbf{Correspondence:} \href{mailto:z.hussain@unibas.ch}{z.hussain@unibas.ch}
 }
}

\begin{document}
\maketitle
\begin{abstract} 
Semantic representations are integral to natural language processing, psycholinguistics, and artificial intelligence. Although often derived from internet text, recent years have seen a rise in the popularity of behavior-based (e.g., free associations) and brain-based (e.g., fMRI) representations, which promise improvements in our ability to measure and model human representations. We carry out the first systematic evaluation of the similarities and differences between semantic representations derived from text, behavior, and brain data. Using representational similarity analysis, we show that word vectors derived from behavior and brain data encode information that differs from their text-derived cousins. Furthermore, drawing on our psychNorms metabase, alongside an interpretability method that we call representational content analysis, we find that, in particular, behavior representations capture unique variance on certain affective, agentic, and socio-moral dimensions. We thus establish behavior as an important complement to text for capturing human representations and behavior. These results are broadly relevant to research aimed at learning human-aligned semantic representations, including work on evaluating and aligning large language models.     
\end{abstract}

\section{Introduction}

Semantic representations are now a staple in computational linguistics \cite{boleda2020distributional}, cognitive psychology \citep{gunther2019vector, hussain2024tutorial}, as well as natural language processing and artificial intelligence \citep{jm3}. Crucial to their success has been the widespread availability of massive corpora of web text \citep[e.g., Common Crawl;][]{commoncrawl} used for representation learning. However, recent years have seen an uptick in novel behavioral and neuroscientific datasets for training and evaluating semantic representations, with the broad hope being that these can capture dimensions of human psychology less present in text. 

For instance, driven by recent large-scale \textit{behavior} data collection efforts \citep[e.g.,][]{de2019small, hebart2023things}, researchers are deriving semantic representations from free associations \citep{richie2021similarity, hussain2024novel, wulff2022semantic}, odd-one-out judgments \citep{hebart2020revealing}, and sensorimotor ratings \citep{kennington2021enriching}. Likewise, semantic representations obtained from \textit{brain} data (e.g., fMRI, EEG) are becoming popular for evaluating and improving the human-likeness of state-of-the-art AI systems: namely, large language models (LLMs) \citep[\href{https://github.com/brain-score/language}{github.com/brain-score/language},][]{hollenstein2019cognival, toneva2019interpreting}. 

\begin{figure*}[t]
  \includegraphics[width=\textwidth]{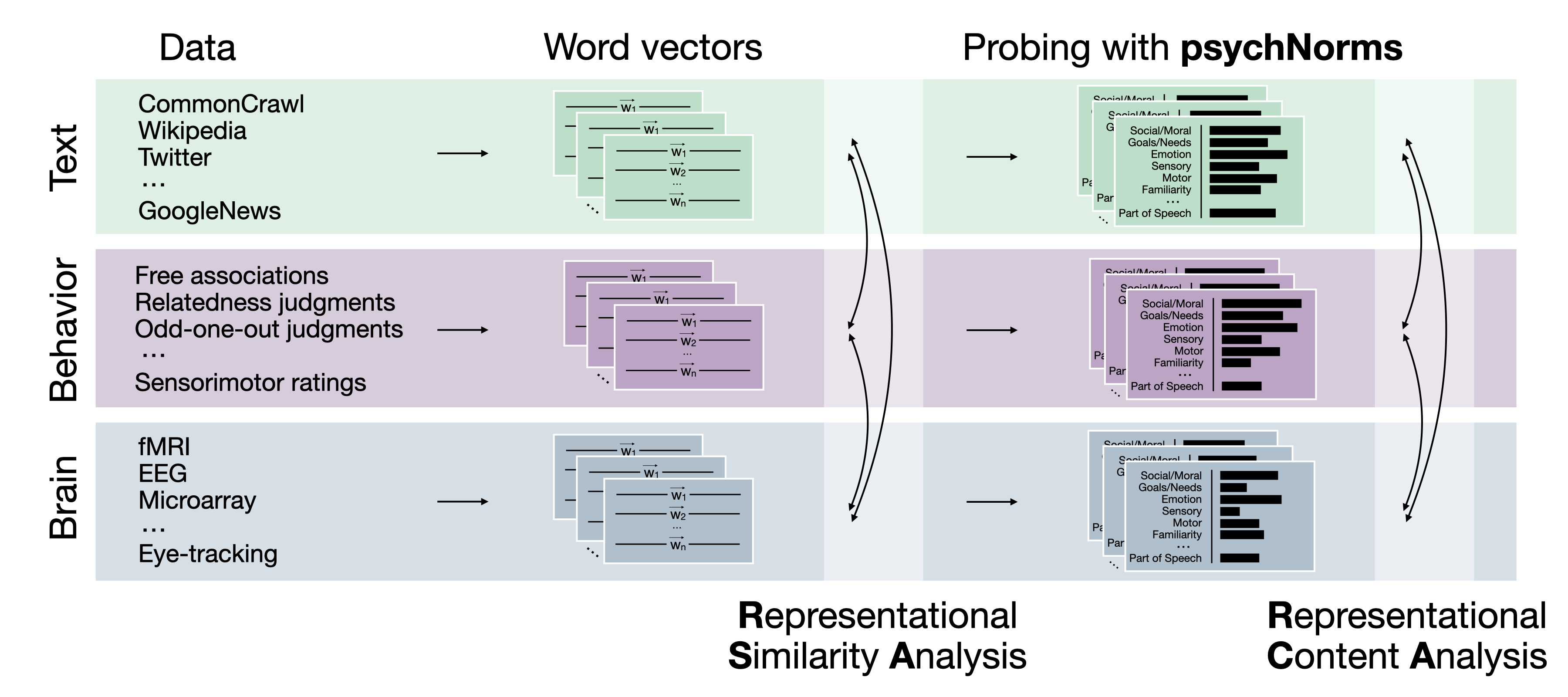}
  \caption{An illustration of our approach. Word vectors are first obtained from the different data sources and then compared via representational similarity analysis (RSA) and representational content analysis (RCA).}
  \label{fig:schema}
\end{figure*}

Our study seeks to address two research questions: (a) do behavior and brain representations encode systematically different information than text, and (b) are these differences relevant from the perspective of measuring and modeling human representations? Figure \ref{fig:schema} illustrates our approach. First, we run a representational similarity analysis (RSA) to uncover systematic differences between text, behavior, and brain data (Section \ref{subsec:rsa_results}). We then analyze the content of these differences via our \textit{representational content analysis} (RCA, Sections \ref{subsec:rca_results}, \ref{subsec:rca_unique}), and end with a discussion of the merits and limitations of our work. 

\section{Our contributions}
\label{sec:contributions}

Our contributions are four-fold. First, we perform a comprehensive comparison of $10$ text representations, $10$ behavior representations, and $6$ brain representations, revealing robust differences between data types (Section \ref{subsec:rsa_results}).

Second, we collate the largest (to our knowledge) metabase of predominantly human-rated (behavioral) word properties (i.e., \textit{word norms}, Section \ref{subsec:reps_norms}), which we call \textit{psychNorms}. The metabase is publicly available at \href{https://github.com/Zak-Hussain/psychNorms}{github.com/Zak-Hussain/psychNorms}, and reflects over half a century of psycholinguistic research. We hope it will serve as a valuable resource for researchers seeking to measure and interpret abstract language representations along psychologically meaningful dimensions. 

Third, leveraging \textit{psychNorms} and linear probes \citep[see, e.g.,][]{belinkov2022probing}, we demonstrate how to build interpretable informational content profiles for abstract representations via a novel analysis framework that we call \textit{representational content analysis} (RCA, Section \ref{subsec:rca_method}). By comparing the profiles of different representations, we can provide crucial insight into the \textit{content} of their differences. This could be especially useful for interpreting and navigating discrepancies between the plethora of otherwise opaque representational alignment metrics \citep{sucholutsky2023getting}. 

Fourth, and most importantly, we show that, despite being trained on orders of magnitudes less data, the behavior representations encode psychological information of comparable and sometimes even superior quality to their text-based cousins (Sections \ref{subsec:rca_results}, \ref{subsec:rca_unique}). This suggests behavior as an important complement to text when it comes to measuring and modeling human representations. 

\section{Methodology}

\subsection{Representations and norms}
\label{subsec:reps_norms}

As mentioned, our analyses seek to answer two questions: (a) do behavior and brain representations encode systematically different information than text, and (b) are these differences relevant from the perspective of measuring and modeling human representations? We attempt to answer these questions using numerical word-level representations (i.e., \textit{word vectors}). These continuous representations permit quantitative comparisons across otherwise incommensurate data types (text, behavior, and brain data). Furthermore, because the representations are at the level of individual words (i.e., there is one vector per word), they can be directly probed using widely available word-level ratings (norms) such as those we collate in \textit{psychNorms}. 
\textit{}
\begin{table*}[t!]
\caption{Text, behavior, and brain representations (*trained as part of this research).}
\label{tab:representations}
\begin{center}
\begin{tabular}{p{0.25\textwidth} p{0.7\textwidth}}
\multicolumn{1}{c}{\bf REPRESENTATION}  & \multicolumn{1}{c}{\bf Description} \\
\hline \\
fastText CommonCrawl & fastText architecture \citep{mikolov2018advances}, trained on CommonCrawl. \\ 
GloVe CommonCrawl & GloVe architecture \citep{pennington2014glove}, trained on CommonCrawl. \\
LexVec CommonCrawl & LexVec architecture \citep{salle2016matrix}, trained on CommonCrawl. \\
fastText Wiki News & fastText architecture \citep{mikolov2018advances}, trained on Wikipedia 2017, UMBC webbase, and statmt.org news. \\
CBOW GoogleNews & CBOW architecture \citep{mikolov2013efficient} trained on the Google News. \\
fastTextSub OpenSub & fastText subword architecture \citep{mikolov2018advances} trained on OpenSubtitles \citep{van2021subs2vec}. \\
GloVe Wikipedia & GloVe architecture \citep{pennington2014glove} trained on Wikipedia 2014. \\
spherical text Wikipedia & Spherical text architecture \citep{meng2019spherical} trained on Wikipedia 2019. \\
GloVe Twitter & GloVe architecture \citep{pennington2014glove} trained on Twitter. \\
morphoNLM & Recurrent neural network architecture fine-tuned on morphological informative examples \citep{luong2013better}. \\ 

\hline

norms sensorimotor & Ratings of 6 perceptual modalities and 5 action effectors \citep{lynott2020lancaster} \\
SGSoftMax[In/Out]put SWOW* & [Cue/Response] vectors from Skip-gram softmax architecture \citep[as in, e.g.,][]{goldberg2014word2vec} trained on SWOW \citep{de2019small}.\\
PPMI SVD SWOW* & Positive pointwise mutual information (PPMI) followed by singular value decomposition (SVD) of the SWOW cue-response matrix \citep[following, e.g.,][]{richie2021similarity, hussain2024novel}. \\
PPMI SVD EAT* & PPMI followed by SVD of the Edinburgh Associative Thesaurus cue-response matrix \citep[EAT,][]{kiss1973associative}. \\
SVD similarity relatedness* & SVD of a similarity matrix of aggregated and normalized similarity and relatedness judgment datasets. \\
feature overlap & Cosine similarity matrix of overlapping feature frequency percentages between cue pairs in a feature listing task \citep{buchanan2019english} \\
THINGS & Neural network embedding trained to predict odd-one-out judgments of image triplets \citep{hebart2020revealing}. \\
experiential attributes & Human ratings on 65 attributes comprising sensory, motor, spatial, temporal, affective, social, and cognitive experiences \citep{binder2016toward} \\ 

\hline

eye tracking & Features extracted from Gaze patterns while reading, aggregated by \citet{hollenstein2019cognival} from 7 datasets.\\
EEG text & Electrode measures while reading sentences \citep{hollenstein2018zuco}.\\
EEG speech & Electrode measures while listening to sentences \citep{broderick2018electrophysiological}. \\
fMRI text hyper align & 1000 randomly-sampled voxels while reading sentences \citep{wehbe2014simultaneously}, processed by \citet{hollenstein2019cognival} and hyper-aligned* across individuals \citep{heusser2017hypertools}.\\
microarray & Neuron-level recordings while listening to sentences \citep{jamali2024semantic}.  \\ 
fMRI speech hyper align & 6 regions of interest while listening to sentences, collected and processed by \citet{brennan2016abstract}, and hyper-aligned* across individuals. \\
\end{tabular}
\end{center}
\end{table*}

Our analyses rely on $10$ text, $10$ behavior, and $6$ brain representations, and $292$ word norms grouped into $27$ norm categories (see Tables \ref{tab:representations} and \ref{tab:norms} for details). For our purposes, we subset each representation to a specific vocabulary. That is, for a given representation \(i\), we compute the intersection of its original vocabulary \(V_i\) with a restricted base vocabulary $V_\text{base}$:

\[
V_i' = V_i \cap V_\text{base},
\]

where  $V_\text{base}$ is defined as the intersection of two components: (a) the union of each and every norm vocabulary \(V_k\), where \(k \in K\) (and \(K\) is the set of all norms), and (b) the union of each and every behavior representation vocabulary \(V_h\) (where \(h \in H\)) and brain representation vocabulary \(V_q\) (where \(q \in Q\)). Formally, this is expressed as:

\[
V_\text{base} = \left( \bigcup_{k \in K} V_k \right) \cap \left( \ \bigcup_{h \in H} V_h \cup \bigcup_{q \in Q} V_q \right).
\]

We do this for three reasons. First, it reduces the most numerous (text) vocabularies to a computationally feasible subset for representational similarity analysis (RSA, Section \ref{subsec:rsa_methods}). Second, it ensures a more controlled comparison between representations by constraining their vocabularies to a more common subset. Finally, it focuses the analyses on a more psychologically relevant set of words---relevant in the sense that they are words that behavioral and neuroscientists have deemed interesting enough for inclusion in their data collection efforts. In concrete terms, $V_\text{base}$ covers $83.6\%$ ($4.62 \times 10^4$ words) of global word occurrences \citep[based on American subtitles][]{brysbaert2009moving}, resulting in a median global coverage of $66.3\%$ and median vocabulary size of $1.18 \times 10^4$ words among the subsetted vocabularies ($V_i'$s).  

\begin{figure*}[t]
  \includegraphics[width=\textwidth]{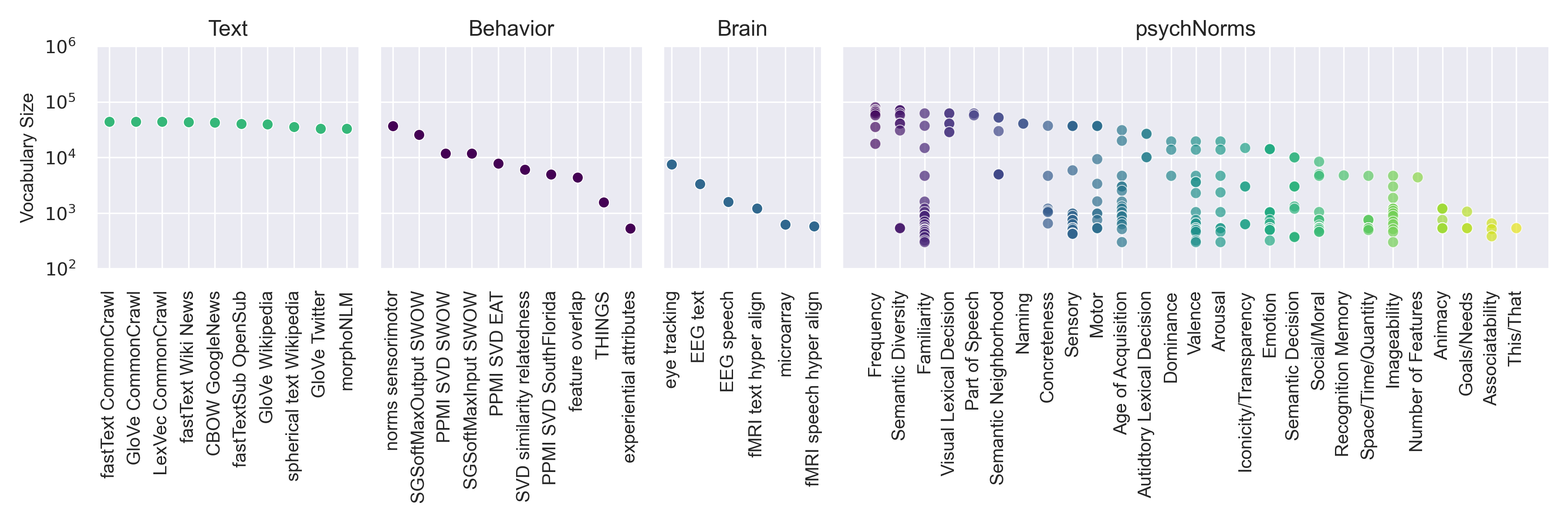}
  \caption{An illustration of the size of the vocabularies (y-axis, log-scaled) for each representation and norm (x-axis, grouped into higher-level categories) used in our analyses. The representations have been grouped into each data type (text, behavior, and brain).}
  \label{fig:vocab_sizes}
\end{figure*}

Figure \ref{fig:vocab_sizes} illustrates the vocabulary sizes in log space. Starting from the left, the text representations possess the largest vocabularies ranging from $3.28 \times 10^4$ to $4.44 \times 10^4$ words (following subsetting). Given text's dominance as a data source for training word representations, we were able to obtain a diverse set of high-quality \textit{pre-trained} representations from publicly available sources (see Table \ref{tab:representations}). 

The behavior representations vary considerably in their vocabulary sizes (min: $5.34 \times 10^2$, max: $3.69 \times 10^4$), with the smallest (\textit{experiential attributes}) on par with the smallest brain representations and the largest (\textit{norms sensorimotor}) approaching that of text. We use a mixture of out-of-the-box behavior representations (e.g., \textit{norms sensorimotor, feature overlap, THINGS}) and those we train ourselves (e.g., \textit{SGSoftMax[Input/Output] SWOW}, \textit{PPMI SVD [SWOW/EAT/SouthFlorida]}). For the latter, we rely heavily on the \textit{Small World of Words} (SWOW) dataset \citep{de2019small}, which is the largest dataset of free associations available. It contains roughly 3.6 million associates to over 12,000 cues, and has been found to be an effective way to uncover semantic representations in humans \citep{aeschbach2024mapping, wulff2022semantic}.

Turning to the brain representations, vocabularies tend to be one or two orders of magnitudes smaller (min: $5.79 \times 10^2$, max: $7.49 \times 10^3$). We draw on preexisting fMRI and EEG data from reading \citep[\textit{[fMRI/EEG] text},][]{hollenstein2018zuco, wehbe2014simultaneously}) and listening \citep[\textit{[fMRI/EEG] speech},][]{broderick2018electrophysiological, brennan2016abstract}) tasks, eye-tracking data from reading tasks \citep[\textit{eye tracking},][]{hollenstein2018zuco}\footnote{Although eye-tracking data is not typically considered brain data, we anticipated that the specific eye-tracking data used in this study, which was obtained from \textit{reading tasks}, would be more closely linked to visual attention than, for instance, semantic relatedness judgments, which we view as more brain-like.}, and a promising novel dataset of neuron-level recordings obtained from tungsten micro-electrode arrays (\textit{microarray}) during listening tasks \citep{jamali2024semantic}. Both EEG representations (\textit{EEG [text/speech]}) and \textit{fMRI text hyper align} were processed by \citet{hollenstein2019cognival} using standard pipelines \citep{beinborn2019robust}, whereas \textit{fMRI speech hyper align} was processed by \citet{brennan2016abstract}.  

\begin{table*}[t!]
\caption{Norm categories (*human-rated/behavioral norms).}
\label{tab:norms}
\begin{center}
\begin{tabular}{p{0.22\textwidth} p{0.68\textwidth}}
\multicolumn{1}{c}{\bf Category}  & \multicolumn{1}{c}{\bf Description} \\
\hline \\
Frequency &  (Log) frequency of word's occurrence in various text corpora. \\
Semantic Diversity & Measures word's polysemy or contextual diversity. \\ 
Familiarity* & Measures how well-known or familiar the word is. \\ 
Visual Lexical Decision* & Measures accuracy or response time during visual decision tasks with the word. \\ 
Part of Speech & The word's dominant grammatical category. \\ 
Semantic Neighborhood* & Network measures of the number and strength of the word's relationships with its neighbors. \\ 
Naming* & Measures accuracy or response time for word naming. \\ 
Concreteness* & Ratings of how concrete or abstract a word is. \\ 
Sensory* & Ratings of how strongly or easily the word is experienced through particular senses. \\ 
Motor* & Ratings of how much a word concerns bodily action or interaction. \\ 
Age of Acquisition* & Estimates of the age at which a word is learned. \\ 
Auditory Lexical Decision* & Measures accuracy or response time during auditory decision tasks with the word. \\ 
Dominance* & Ratings of the degree to which the word can be controlled. \\ 
Valence* & Ratings of how positive or negative a word is. \\ 
Arousal* & Ratings of the intensity of emotion or excitation evoked by a word. \\ 
Iconicity/Transparency* & Ratings of how much a word looks or sounds like what it means. \\ 
Emotion* & Ratings of how much a word reflect or elicits certain emotions. \\ 
Semantic Decision* & Accuracy or response time during semantic rating tasks. \\ 
Social/Moral* & Ratings of a word's relevance to social and moral dimensions.  \\ 
Recognition Memory* & Recognition memory performance (hits minus false alarms). \\ 
Space/Time/Quantity* & Ratings of a word on spatial, temporal, and other quantitative dimensions. \\ 
Imageability* & Ratings of the ease with which a word can be imagined. \\ 
Number of Features* & Number of features listed for a word. \\ 
Animacy* & Ratings of how much a word is thinking, living, or human-like. \\ 
Goals/Needs* & Ratings of how much a word represents goals, needs, or drives. \\ 
Associatability* & Ratings of how quick and easy it is to thing of associations to a word. \\ 
This/That* & Proportion of times participants associated words with \textit{this} versus \textit{that}. \\
\end{tabular}
\end{center}
\end{table*}

Finally, in order to measure the psychological content of the representations (via RCA), we needed a vast dataset of existing norms. Although norm (meta-)databases exist \citep[e.g.,][]{gao2023scope}, there are (to our knowledge) no systematic literature searches for human-rated word properties. We thus screened 3,056 articles containing norm-relevant keywords (returning 181 norms) and combined the results with selected norms from the largest preexisting norm metabase \citep[SCOPE, 97 norms selected,][]{gao2023scope} and a dataset of 65 human-rated experiential attributes \citep{binder2016toward}. This resulted in a metabase of 292 unique norms, which we call psychNorms and make available at \href{https://github.com/Zak-Hussain/psychNorms}{github.com/Zak-Hussain/psychNorm}.  

As illustrated on the right-hand side of Figure \ref{fig:vocab_sizes}, these norms differ considerably both in the size of their vocabularies and the kinds of properties they seek to measure. To aid in the interpretation of this diversity, we have manually grouped the norms (points) into higher-level categories (x-axis) (see Table \ref{tab:norms}). These categories include those that are popular in natural language processing settings (e.g., Frequency, Part of Speech, and Valence) as well as categories that have hitherto been relatively constrained to psycholinguistics (e.g., Space/Time/Quantity, Animacy, Goals/Needs). 

\subsection{Representational similarity analysis}
\label{subsec:rsa_methods}

We use representational similarity analysis (RSA) to compare the information encoded in the above representations. Developed within neuroscience \citep{kriegeskorte2008representational}, RSA enables comparisons of representations from otherwise-disparate modalities (e.g., fMRI, EEG, similarity ratings) by leveraging the fact that the different dimensions may nevertheless contain information that seeks to distinguish a comparable set of mental states, stimuli, or other kinds of entities. 

In our case, the entities being distinguished are words. Consequently, RSA measures the similarity between two matrices, \(\mathbf{M}_i\) and \(\mathbf{M}_j\), where each row represents a word, and each column reflects a measurement unit (dimensions). For the brain representations, these units may be voxels (fMRI) or electrode readings (EEG), whereas, for text and behavior models, the units are often abstract dimensions. RSA addresses the challenge of correlating these different units by transforming \(\mathbf{M}_i\) and \(\mathbf{M}_j\) into a common space. This transformation is achieved by calculating the (dis)similarities between the rows of \(\mathbf{M}_i\) and \(\mathbf{M}_j\), forming what is known as a \textit{representational similarity matrix}, $\mathbf{S}$. Following \citet[e.g.,][]{lenci2022comparative}, we compute the \textit{cosine} similarity matrices \(\mathbf{S}_i\) and \(\mathbf{S}_j\), as:

\[
\mathbf{S}_i = \mathbf{\tilde{M}}_i \cdot \mathbf{\tilde{M}}_i^\top \quad \text{and} \quad \mathbf{S}_j = \mathbf{\tilde{M}}_j \cdot \mathbf{\tilde{M}}_j^\top,
\]

\noindent where the tilde notation $\mathbf{\tilde{M}}$ indicates that the rows of the matrices have been \(L_2\) normalized. We then compute the representational similarity \(\rho_{ij}\) between the two representations as the Spearman correlation between \(\mathbf{S}_i\) and \(\mathbf{S}_j\), having subsetted the two matrices to their common vocabularies \(V_{ij}^c = V_i' \cap V_j'\):     

\[
\rho_{ij} = cor(vec_{\operatorname{triu}}(\mathbf{S}_{i,\in V_{ij}^c}), vec_{\operatorname{triu}}(\mathbf{S}_{j,\in V_{ij}^c})),
\]

where $vec_{\operatorname{triu}}(\mathbf{S})$ denotes the vector formed by flattening the upper triangle of $\mathbf{S}$, excluding the diagonal. 

\subsection{Representational content analysis}
\label{subsec:rca_method}

Representational content analysis (RCA) is an approach to interpretable informational content \textit{profiles} for abstract numerical representations. Although it leverages the well-established technique of probing from deep learning interpretability \citep[see, e.g.,][]{belinkov2022probing}, it differs from traditional probing applications \citep[e.g.,][]{conneau2018you, sahin-etal-2020-linspector} in scope and focus. RCA employs tens or even hundreds (as in our case) of targets to more holistically interpret the information encoded in the representations and focuses on revealing how representations derived from different types of data (or learning algorithms) differ from one another.

Our RCA implementation uses L2-regularized linear probing classifiers and regressors to predict each norm (target) from each representation (features). That is, for each representation $i$ and norm $k$, we fit a linear mapping \(\mathbf{w}_{ik}\) such that:

\[
\mathbf{\hat{y}}_{ik,\in V_{ik}^c} = \mathbf{M}_{i, \in V_{ik}^c} \mathbf{w}_{ik},
\]

\noindent where $\mathbf{M}_{i, \in V_{ik}^c}$ is a matrix of word vectors restricted to a common vocabulary \(V_{ik}^c = V_i' \cap V_k\) and $\mathbf{\hat{y}}_{ik, \in V_{ik}^c}$ is a vector of predicted norm ratings. We employ L2-regularization to mitigate issues such as multicollinearity, underdetermination, and over-fitting in high-dimensional settings. Following \citet{hupkes2018visualisation}, we use \textit{linear} probes to avoid the risk of more flexible estimators learning features that do not faithfully reflect what is present in the original representations. 

For numerical norms, we use the Scikit-Learn API's \verb|RidgeCV| \citep{scikit-learn}. For binary and multi-class norms, we use the API's \verb|LogisticRegressionCV|. Both estimators perform automatic (hyperparameter) tuning of the L2 penalty. This parameter---\verb|alpha| in the case of \verb|RidgeCV|, or \verb|C| in the case of \verb|LogisticRegressionCV| (equivalent to \verb|1/apha|)---is selected from a grid of values ranging from $10^{-5}$ to $10^5$ (in \verb|alpha| terms) with even spacing in log (base-10) space. 

To evaluate the probing loss, we calculate the (pseudo) $R^2$ via 5-fold nested cross-validation \citep{scikit-learn}, where the regression coefficients and L2 penalty parameter are fitted in an inner loop, and evaluated on separate test sets in the outer loop \citep[following, e.g.,][]{varma2006bias}. We thus define the measured content $\boldsymbol{\psi}$ for each representation $i$ as a vector of (pseudo) $R^2$ scores across all $n$ norms: 

\[
\boldsymbol{\psi}_i = \left[ R^2_{i1}, R^2_{i2}, \dots, R^2_{in} \right].
\]

Finally, to ensure some minimum reliability for performance estimates, we do not probe cases where the intersection of the representation and norm vocabulary $V_{ik}^c$ results in a test set with fewer than 20 samples. This is important to keep in mind for Section \ref{subsec:rca_results}, where, in a minority of cases, average performances on a given norm category are estimated from a reduced set of norms. 

\section{Experiments}

\subsection{Representations from text, behavior, and brain differ systematically, irrespective of learning algorithm}
\label{subsec:rsa_results}

\begin{figure*}[t]
  \includegraphics[width=\textwidth]{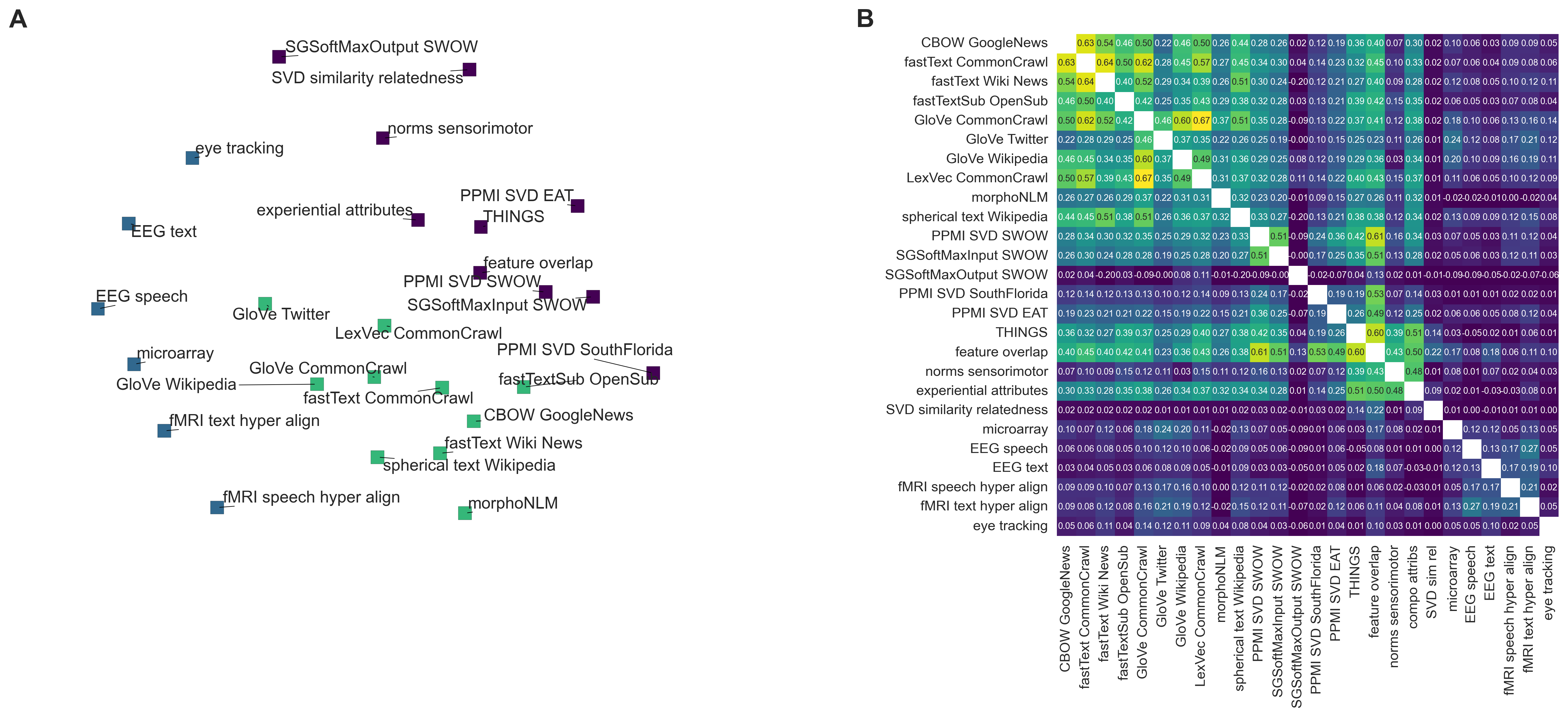}
  \caption{\textbf{A}: A 2-dimensional projection of the representational similarity space. The space was obtained by multidimensional scaling of the pairwise Spearman dissimilarity matrix between representations. Text = green, behavior = purple, brain = blue. \textbf{B}: A heatmap visualization of the pairwise Spearman similarity matrix.}
  \label{fig:rsa}
\end{figure*}

We begin by asking to what extent text, behavior, and brain data encode distinct information. Using representational similarity analysis (RSA), we compare the representations obtained from each data type (see Section \ref{subsec:rsa_methods} for details).

Figure \ref{fig:rsa} illustrates the results. Panel A presents a multidimensional scaling of the representational similarity space, and Panel B presents the pairwise similarity matrix. It is important to emphasize that each data type encompasses a diverse set of representations derived from different learning algorithms and sub-data-types (or sub-datasets) (see Section \ref{subsec:reps_norms} for details). For instance, the text and behavior representations result from algorithms both from the \textit{global matrix factorization} family (e.g., \textit{PPMI SVD} SWOW, \textit{SVD} Similarity Relatedness), \textit{local context window} family (e.g., \textit{fastText} CommonCrawl, \textit{SGSoftMax Input} SWOW), and hybrids of both families (e.g., \textit{GloVe} CommonCrawl).  

Despite the diversity within data type and some algorithmic commonalities between types (e.g., \textit{fastText CommonCrawl}, \textit{SGSoftMax Input SWOW}), we observe very clear clustering by data type (Figure \ref{fig:rsa}) and only mild clustering based on the representation learning algorithm. This suggests that the data type has a more significant effect on representational similarity than the choice of learning algorithm.

To answer our first research question, we find considerable differences between brain and behavior compared to text (text-brain $\bar{\rho}=.09$, text-behavior $\bar{\rho}=.20$, where $\bar{\rho}$ denotes the mean Spearman correlation), with similarities between the data types showing lower values than those within (brain-brain $\bar{\rho}=.12$, behavior-behavior $\bar{\rho}=.22$, text-text $\bar{\rho}=.41$). Moreover, we observe proportions of top-3 nearest neighbors being of the same data type of $.97\%$ (text), $93\%$ (behavior), and $50\%$ (brain), suggesting high affinity among representations of the same data type, especially for text and behavior. Interestingly, the similarity between text and brain turns out to be $.06$ points higher than that between brain and behavior (brain-behavior $\bar{\rho}=.03$), suggesting that behavior is the most distinct data type.  

Ultimately, our analyses demonstrate the importance of data type in shaping representational similarity, with noticeable informational differences between text, behavior, and brain. We now move to characterizing these differences using representational content analysis. 

\subsection{Behavior representations can rival text in psychological content}
\label{subsec:rca_results}

\begin{figure*}[t]
  \includegraphics[width=\textwidth]{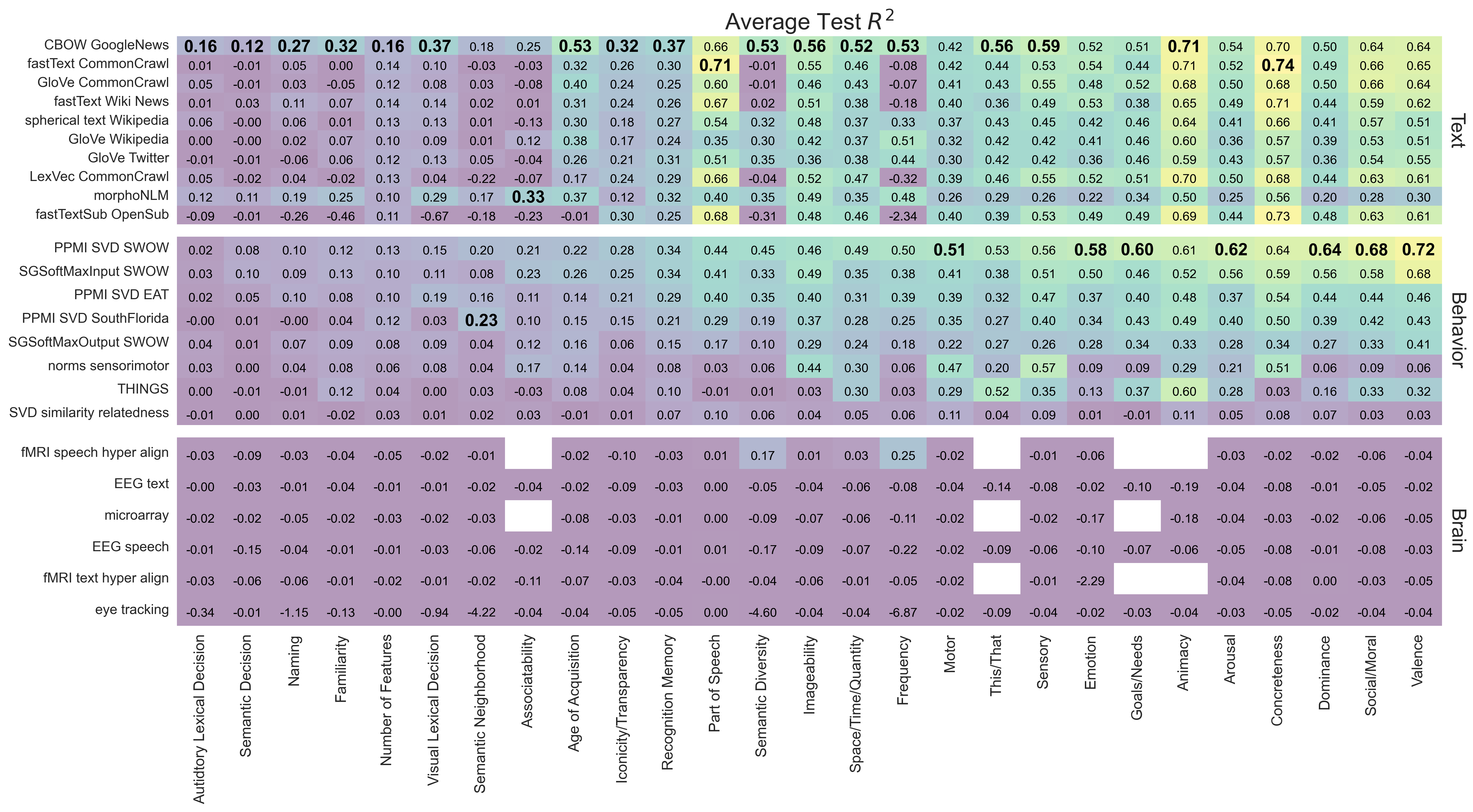}
  \caption{Average 5-fold cross-validation (pseudo-)$R^2$ test performance for text, behavior, and brain representations (rows, grouped) on 292 norms grouped into 27 norm categories (columns). Performances are aggregated by first taking the mean $R^2$ on each norm and then the median of the norm-wise (mean) $R^2$s for each norm category. Representations are ordered within each data type in terms of overall performance. Norm categories are ordered in terms of the performance of the top-performing behavior representation (\textit{PPMI SVD SWOW}). Missing values are the result of an insufficient number of test samples.}
  \label{fig:rca}
\end{figure*}

The previous section revealed differences in the information encoded in text, behavior, and brain representations. This raises the question: What is the \textit{content} of these differences, and are they relevant to measuring and modeling human representations? To address this question, we leverage our \textit{psychNorms} metabase (Section \ref{subsec:reps_norms}) as targets in a representational content analysis (RCA, Section \ref{subsec:rca_method}). 

Figure \ref{fig:rca} illustrates the average test performances of each representation\footnote{\textit{feature overlap} and \textit{experiential attributes} are dropped from remaining analyses due to, respectively, a vast number of missing values (words with no overlapping features were set to NaN), and an insufficiently large vocabulary.} (rows) on each norm category (columns). Performance is measured via the coefficient of determination ($R^2$) for numerical norms and McFadden's pseudo-$R^2$ for categorical norms (e.g., \textit{This/That}, \textit{Part of Speech} norms). We henceforth denote both measures with $R^2$.  

Some interesting patterns can be observed. First, text and behavior appear to encode a broad range of psychological information, with the grand median performance of text ($\tilde{R^2}=.38$, IQR: $.36$-$.41$) considerably higher than that of behavior ($\tilde{R^2}=.22$, IQR: $.08$-$.36$). The strong performance of text is perhaps unsurprising---it has, after all, been the dominant data source for training semantic representations for decades \citep{gunther2019vector}. Behavior, on the other hand, has garnered comparatively little attention. The representations are also derived from orders of magnitudes smaller training sets and possess more modest vocabularies (hence, smaller probe-training sets). Behavior's performance relative to text is thus quite impressive.

Second, we detect scarce psychological information in brain ($\tilde{R^2}=-.04$, IQR: $-.04$-$-.03$). However, it is important to reiterate brain's limited vocabularies here. Furthermore, in many cases, the number of features (e.g., voxels, electrode readings) approaches the number of norm-labeled words (samples), making it all the more difficult to detect norm signals in the brain data, which is anyhow inherently quite noisy \citep{raichle2010two}. Nevertheless, in its present form, brain does not appear to capture much psychological content.

Third, it appears that some norms are in general better-encoded than others across representations: namely, those on the right-hand side (e.g., \textit{Valence}, \textit{Social/Moral}, \textit{Dominance}) of Figure \ref{fig:rca} versus those on the left (e.g. \textit{Auditory Lexical Decision}, \textit{Semantic Decision}, \textit{Naming}). This may be explained in part by differences in norm measurement reliability, which sets a theoretical upper bound on the norm signal that can be captured. However, it is also possible that certain norm-relevant information is especially hard to capture for \textit{all} kinds of semantic representation (perhaps because the information is not semantic in nature). This latter explanation could indicate an avenue for future research seeking to capture remaining psychological information. 

Fourth and finally, important differences can be observed between the best-performing representations from each type on certain norms. For instance, the best-performing text representations tend to outperform those of behavior by a considerable margin on \textit{Age of Acquisition} (absolute difference in maximum $R^2$, $|\Delta R^2_{\text{max}}| = .27$), \textit{Part of Speech} ($|\Delta R^2_{\text{max}}| = .27$), \textit{Familiarity} ($|\Delta R^2_{\text{max}}| = .19$, \textit{Visual Lexical Decision} ($|\Delta R^2_{\text{max}}| = .18$), and \textit{Naming} ($|\Delta R^2_{\text{max}}| = .16$) norms. Of course, these superior performances may be (partially) attributable to the text representations' larger vocabularies and pre-training sets (we control for probe-training set size and constitution in the next section, \ref{subsec:rca_unique}). The differences are nevertheless notable. 

Conversely, the best-performing behavior representations perform comparatively strongly on \textit{Dominance} ($|\Delta R^2_{\text{max}}| = .15$), \textit{Motor} ($|\Delta R^2_{\text{max}}| = .09$), \textit{Goals/Needs} ($|\Delta R^2_{\text{max}}| = .08$),  \textit{Arousal} ($|\Delta R^2_{\text{max}}| = .08$), and \textit{Valence} ($|\Delta R^2_{\text{max}}| = .07$) norms, relative to text. Given the behavior representations' smaller vocabularies and pre-training sets, these higher performances can be seen as conservative estimates of the content behavior is capturing and what it can contribute beyond text to modeling human representations. 

All in all, our RCA provides critical insights into the content of the differences between text, behavior, and brain representations. Having identified a surprisingly rich reservoir of psychological information in behavior, we now move on to the question of the extent to which behavior could complement text when it comes to modeling human representations.   

\subsection{Behavior representations captures unique psychological variance}
\label{subsec:rca_unique}

\begin{figure*}[t]
  \includegraphics[width=\textwidth]{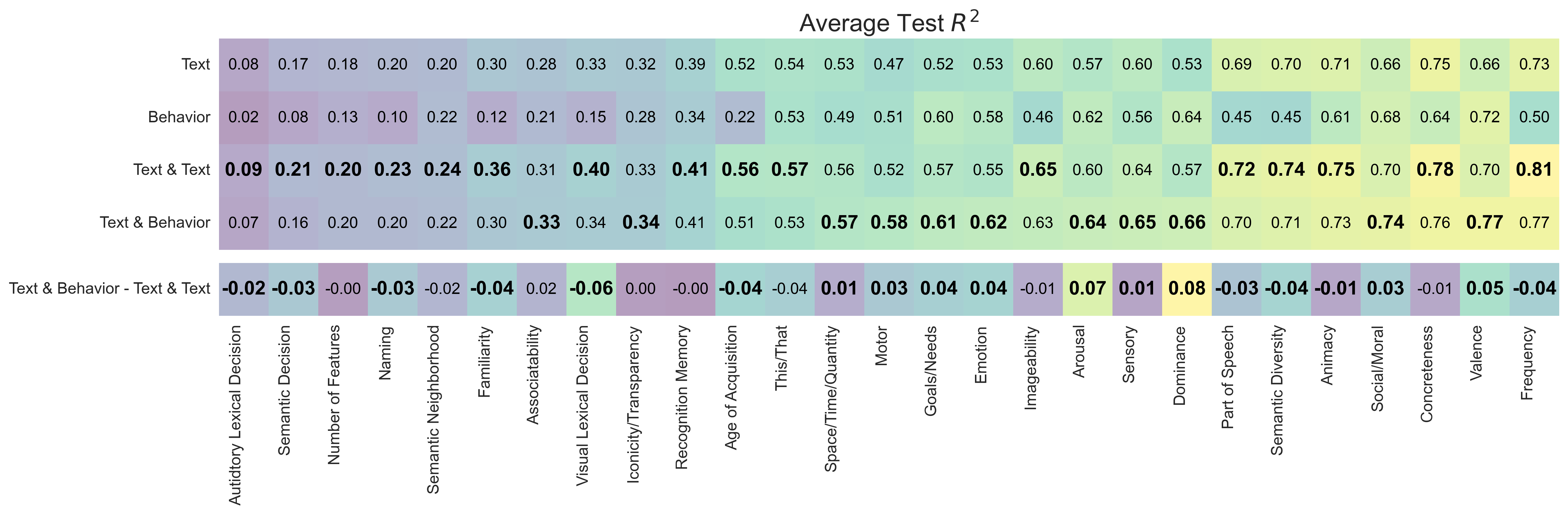}
  \caption{Average 5-fold cross-validation (pseudo-)$R^2$ performance for the top-2 \textit{Text} (\textit{CBOW GoogleNews} and \textit{fastText CommonCrawl}), top \textit{Behavior} (\textit{PPMI SVD SWOW}) representations, and all \textit{Text \& Text} and \textit{Text \& Behavior} ensemble combinations. Performances are aggregated by first taking the mean (difference in) $R^2$ on each norm and then the median of the norm-wise (mean) $R^2$ for each norm category. Norms are ordered in terms of the performance of \textit{Text \& Behavior}. Emboldened differences (\textit{Text \& Behavior -  Text \& Text}) reflect a Wilcoxon signed rank $p<.05$.}   
  \label{fig:rca_ensemb}
\end{figure*}

The previous section suggests that behavior representations contain psychological information that text fails to capture. We now turn to the question of the \textit{unique} (marginal) contribution of behavior beyond text. To investigate this, we perform an ensemble RCA, whereby we concatenate the top-performing text and behavior representations and measure the marginal increase in norm variance explained. This approach allows us to disentangle the unique norm variance captured by adding behavior features while holding the text features constant. We also subset all representation vocabularies to their collective intersection, meaning that the size and content of the probe's training set on any given norm are identical across representations, thus improving comparability between representations. 

Figure \ref{fig:rca_ensemb} illustrates the results. We contrast text and behavior ensembles (\textit{Text \& Behavior}) with \textit{Text \& Text} and provide solo \textit{Text} and \textit{Behavior} baselines for reference. The first thing to note is that ensembling tends to improve performance: on any given norm, it is either \textit{Text \& Text} or \textit{Text \& Behavior} in first place. 

However, neither \textit{Text \& Text} nor \textit{Text \& Behavior} is the unanimous winner. For instance, consistent with the results in in Section \ref{subsec:rca_results}, \textit{Text \& Text} tends to outperform \textit{Text \& Behavior} on \textit{Visual Lexical Decision} (absolute median difference, $|\tilde{d}|=.06$), frequency-related norms ($|\tilde{d}|=.04$ for \textit{Age of Acquisition}, \textit{Familiarity}, and \textit{Frequency}), and \textit{Semantic Diversity} ($|\tilde{d}|=.04$), with all mentioned differences having a Wilxocon signed rank $p<.05$ (as indicated by the emboldened differences in Figure \ref{fig:rca_ensemb}). 

\textit{Text \& Behavior}, on the other hand, tends to perform better on affect-related norms (\textit{Dominance}: $|\tilde{d}|=.08$, \textit{Arousal}: $|\tilde{d}|=.07$, \textit{Valence} $|\tilde{d}|=.05$, \textit{Emotion}: $|\tilde{d}|=.04$), agency-related norms (\textit{Goals/Needs}: $|\tilde{d}|=.04$, \textit{Motor}: $|\tilde{d}|=.03$), and \textit{Social/Moral} ($|\tilde{d}|=.03$) norms (all with $p<.05$). 

Ultimately \textit{Text \& Behavior} (descriptively) outperforms \textit{Text \& Text} on 11 out of the 27 norm categories. While \textit{Text \& Text} tends to perform better on more objective, frequency-based categories, \textit{Text \& Behavior} performs better on more subjective categories (e.g., affective, agential, \textit{Social/Moral}). These subjective categories are thought to be fundamental components of human representations \cite[see, e.g.,][]{lynott2020lancaster, binder2016toward}, and are thus likely to be important for language representation applications that seek to model human representations and behavior, including sentiment analysis \citep{socher2013recursive}, cognitive modeling \citep{gunther2019vector}, in-silicon behavioral experiments \citep{yax2024studying}, and AI assistants \citep{bai2022training}.  

\section{Discussion}

This article began by asking whether behavior and brain representations might differ from text representations in ways that are relevant to measuring and modeling human representations. We showed that behavior and brain representations encode information that differs from text representations (Section \ref{subsec:rsa_results}). Drawing on our \textit{psychNorms} metabase and representational content analysis (RCA), we probed these representations to reveal rich, interpretable psychological profiles, with behavior outperforming text on several dimensions (e.g., \textit{Dominance}, \textit{Arousal}, Section \ref{subsec:rca_results}). Motivated by evidence suggesting psychologically important differences between text and behavior, we carried out an ensemble analysis to reveal significant improvements from ensembling text with behavior on affective, agentic, and \textit{Social/Moral} dimensions. We view our work as the first comprehensive analysis of the contents of these different data types, revealing important ways in which they---namely, text and behavior---can complement each other. 

Our findings have important implications. First, as mentioned, many applications of semantic representations (e.g., sentiment analysis, cognitive modeling) attempt to measure or model human representations and behavior \citep{wulff2022semantic, aeschbach2024mapping, gunther2019vector}. Our results indicate that using behavior or supplementing text representations with behavior could improve the quality of these applications, especially insofar as they involve affective, agentic, or \textit{Social/Moral} components (which are considered fundamental to cognition). Second, our findings suggest that the relative performance of the behavior representations could be significantly improved with greater investment in behavior datasets, which are presently several orders of magnitude smaller than those used for pre-training text representations. This investment could be especially important for our third implication, which concerns the measurement and improvement of human-LLM alignment. On the \textit{measurement} front, our RSA and RCA findings can be used to better understand the contents of datasets already used in the evaluation of language models, such as textual similarity judgments \citep[e.g.,][dataset]{cer2017semeval}, sentiment judgments \citep[e.g.,][dataset]{socher2013recursive}, brain imaging data \citep{hollenstein2019cognival} and, more prospectively, free associations \citep{thawani2019swow, abramski2024llm}. Regarding \textit{improving} human-LLM alignment, consistent with the current practice of pre-training on text and fine-tuning on human behavior \citep{bai2022training}, our findings suggest that LLMs that are trained on additional sources of behavior data (e.g., free associations) could become better-aligned on certain psychological dimensions.

Our work has some limitations. First, our approach does not allow for perfectly controlled content comparisons between representations. As mentioned in Section \ref{subsec:rca_results}, although better probing results \textit{may} signal the encoding of more norm-relevant information, they may also reflect larger probe-training set sizes. These issues can be alleviated by subsetting to the same vocabulary across comparison conditions (as we do in Section \ref{subsec:rca_unique}). However, this will naturally reduce the probe's sensitivity to norm-relevant signals due to decreased training set size from subsetting. 

Another limitation concerns our brain data, where we detect little evidence of psychological information. Although this may simply be due to the brain representations' small vocabularies and the inherent noisiness of brain data, it could also be that brain data is poorly suited to word-level analyses such as ours. After all, the brain data was collected during sentence-level tasks, meaning word-level representations had to be extracted via relatively crude heuristics (e.g., a four-second hemodynamic delay offset) and averaging across contexts \citep{hollenstein2019cognival}. We would thus caution against drawing strong conclusions against other brain data formats (e.g., \href{https://github.com/brain-score/language}{github.com/brain-score/language}) on these bases. Nevertheless, the brain representations included in our analysis are taken from recent work seeking to measure and understand human semantic representations \citep[e.g.,][]{hollenstein2019cognival, brennan2016abstract, jamali2024semantic}, suggesting that the brain representations reflect the current state in the literature in terms of word-level representations.   

\section{Conclusion}

In this work, we compared behavior and brain representations with text representations on a broad set of psychological dimensions. We found that, despite limited training data, behavior captures psychological information sometimes rivaling that of text, and also captures unique psychological variance on certain dimensions. Our work thus establishes behavior representations as important complements to their text-based cousins in measuring and modeling human semantic representations. 

\section{Acknowledgments}

We thank Laura Wiles for editing the manuscript. This work was supported by grants from the Swiss Science Foundation to Rui Mata (204700) and Dirk U. Wulff (197315).

\section{Code and data availability}

Code and data for reproducing the analyses in this paper are available at \href{https://github.com/Zak-Hussain/psychProbing}{github.com/Zak-Hussain/psychProbing}. More information on the \textit{psychNorms} metabse, including specifics of the literature search and norm metadata, can be found at \href{https://github.com/Zak-Hussain/psychNorms}{github.com/Zak-Hussain/psychNorms}. 

\bibliography{custom}


\end{document}